%% file: main.tex
\documentclass[runningheads]{llncs}

\usepackage{eccv}

\usepackage{eccvabbrv}
\usepackage{amsmath}
\usepackage{bbm}

\usepackage{pifont}
\usepackage{booktabs}

\usepackage[pagebackref,breaklinks,colorlinks]{hyperref}

\usepackage{orcidlink}

\usepackage{colortbl}
\usepackage{makecell}

\begin{document}

\title{Depth-guided NeRF Training via Earth Mover’s Distance} 

\titlerunning{Depth-guided NeRF Training via Earth Mover’s Distance}

\author{
Anita Rau\and
Josiah Aklilu\and
F. Christopher Holsinger\and
Serena Yeung-Levy
}

\authorrunning{A. Rau et al.}

\institute{
Stanford University \\
\email{\{arau, josaklil, holsinger, syyeung\}@stanford.edu} \\
\url{https://anitarau.github.io/emd-nerf.github.io/}
}

\maketitle

\begin{abstract}
Neural Radiance Fields (NeRFs) are trained to minimize the rendering loss of predicted viewpoints. However, the photometric loss often does not provide enough information to disambiguate between different possible geometries yielding the same image. Previous work has thus incorporated depth supervision during NeRF training, leveraging dense predictions from pre-trained depth networks as pseudo-ground truth. While these depth priors are assumed to be perfect once filtered for noise, in practice, their accuracy is more challenging to capture. This work proposes a novel approach to uncertainty in depth priors for NeRF supervision. Instead of using custom-trained depth or uncertainty priors, we use off-the-shelf pretrained diffusion models to predict depth and capture uncertainty during the denoising process. Because we know that depth priors are prone to errors, we propose to supervise the ray termination distance distribution with Earth Mover's Distance instead of enforcing the rendered depth to replicate the depth prior exactly through $L_2$-loss. Our depth-guided NeRF outperforms all baselines on standard depth metrics by a large margin while maintaining performance on photometric measures.
\end{abstract} 

\section{Introduction}
\label{sec:intr}

Neural Radiance Fields (NeRFs) \cite{nerf} have demonstrated an impressive ability to render novel views of a known scene. Especially in object-centric and well-sampled scenes, NeRFs can generate photometrically and geometrically consistent images from previously unseen view points. However, in camera-centric and sparse view scenarios, neural radiance fields have yet to show the same fidelity. Additionally, while producing renderings of high quality, some NeRFs fail to capture the underlying geometry of a scene accurately, which is essential for applications in robotics or augmented reality \cite{wang2022neural,blukis2023one,adamkiewicz2022vision,zhu2022nice}. Some reasons include a smaller overlap between images, occlusions between views, and photometric inconsistencies due to the camera's auto exposure. 

To improve the robustness of NeRFs in complex indoor settings, previous work \cite{darf, scade, dsnerf, densedepthprior} has incorporated depth supervision during training. The idea is that a better understanding of the underlying scene geometry will enhance image rendering. However, dense depth ground truths are rarely available. 
\begin{figure}[t]
\footnotesize
    \includegraphics[width=\linewidth]{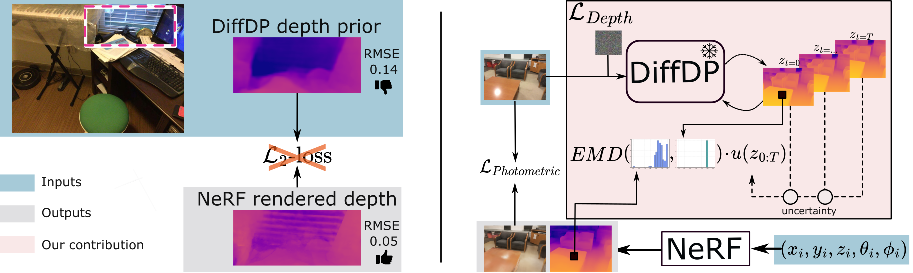}
    \caption{    Left: Predicted monocular depth priors are not perfect and false interpretations of a scene's geometry are unavoidable. Blindly forcing a NeRF to replicate such priors (e.g. through $L_2$-loss) leads to high geometric losses. Right: Overview of our method. We use depth priors to guide NeRF training via Earth Mover's Distance (EMD).}
    \label{fig:pull}
\end{figure}

Fortunately, a line of work has shown that depth estimation networks have become accurate at predicting the layout of indoor environments \cite{ranftl2020towards,zoedepth, dinov2, ddp,ranftl2021vision} and they can be used as pseudo ground truth for NeRF training. But when used ineffectively, depth supervision can fail to improve the NeRF's depth renderings, suggesting a deficient understanding of scene geometry. One reason is that, although depth prediction with neural networks has improved dramatically, there is still ambiguity owed to the ill-defined nature of estimating depth from one view, as many geometric layouts can lead to the same image. Partial occlusion of objects, shading, and reflections further challenge monocular depth prediction, leading to misinterpreted geometries. 

Several methods have addressed the described ambiguity in monocular depth. Some works assume that, once a depth prior is accepted, all pixels within an image are strong pseudo ground truths. \cite{scade} proposed to incorporate expensive, custom-trained multi-modally distributed depth hypotheses of which one is picked for supervision at each training step. However, we observe that generating additional depth predictions comes at the price of generating harmful noise. \cite{darf} treats patches individually, but assumes that all pixels in a patch are either a perfect ground truth or not useful at all, when in practice all pixels have varying levels of uncertainty. Other methods do consider uncertainty at the pixel-level, but treat depth priors as normally distributed around the true depth, which is over-simplified and empirically does not hold \cite{densedepthprior, dsnerf}. To our knowledge, none of the previous methods have captured notions of uncertainty in depth priors and leveraged them for NeRF training effectively.

We argue that although monocular depth predictions can be highly inaccurate, some are useful---so it is crucial to avoid a strict adherence to the depth prior in NeRF's predictions. Depth priors should be a \textit{suggestion}. We propose to supervise the ray termination distance of a NeRF with the Earth Mover's Distance ($EMD$) as illustrated in Figure \ref{fig:pull}. This invites the NeRF to sample ray termination distances close to the depth prior, without directly competing for the final weighting of ray termination distances with the photometric loss. Unlike previous works, we avoid $L_2$-losses that will, inevitably, enforce incorrect depth as well, such as the missed window blinds in Figure \ref{fig:pull}. Instead, EMD allows us to maintain useful information about the distribution of the ray termination distances, while avoiding restrictive assumptions such as normally distributed errors, unimodality, continuity, or non-zero probabilities such as KL-divergence. 

Our work employs out-of-the-box pre-trained generative diffusion models as depth priors and leverages the denoising process for additional uncertainty estimation. This provides accurate depth priors and uncertainties for free. These uncertainty maps tell us when the depth model is unsure, and when we should rely more on the RGB loss. We then introduce an approach to weighing our novel depth loss and the photometric loss that is inspired by Focal Loss \cite{lin2017focal}.

In summary, we propose a new way to think about uncertainty in depth supervised NeRF. We provide novel techniques to incorporate pixel-wise uncertainty, without imposing restrictive assumption on the nature or distribution of the uncertainty, the depth prior, or NeRF ray termination distances. We outperform all baselines on all depth metrics by at least 11\% on ScanNet \cite{dai2017scannet}, and outperform the most closely related baseline by up to 54\% on the relative error. The results speak to our method's ability to understand the underlying geometry of a scene rather than just rendering accurate looking images based upon an incorrect understanding of the 3D world. 

\section{Related Work}
\label{sec:relatedwork}

\paragraph{Monocular depth estimation:}
 Depth estimation from a single view is inherently ambiguous but many works have achieved impressive depth accuracy \cite{li2023depthformer,newcrfs,zoedepth,ddp,depthgen,metric3d,depthanything}. \cite{zoedepth} describes a method for depth estimation that generalizes well to multi-domain data by bridging the gap between relative and metric depth estimation. \cite{metric3d} uses a large-scale dataset and canonical camera transformations on input views to resolve the same metric ambiguity in zero-shot. \cite{dinov2, swinmim, depthanything} demonstrate how rich representations from self-supervised pretraining enables strong performance in dense visual perception. Most recently, \cite{ddp, depthgen} leverage image-conditioned denoising diffusion in the depth estimation pipeline. In our work, we capitalize on the robust uncertainty estimates readily available in generative approaches to depth prediction. We use the monocular depth estimation network DiffDP \footnote{Both \cite{ddp} and \cite{densedepthprior} are called DDP, so we refer to them as DiffDP and DDPrior.} \cite{ddp} to provide depth pseudo ground truths for NeRF.

\paragraph{NeRF with sparse views:}
Neural radiance fields \cite{nerf} have become a popular choice for 3D scene representations due to their ability to render accurate novel views. NeRF comprises of a neural network that takes as input coordinates and camera parameters and outputs color and density. However, the original NeRF relies crucially on many inputs views from the scene of interest in order to faithfully reconstruct the scene geometry. Additionally, it is known that naively training NeRF tends to oversample empty space in the scene volume, leading to artifacts and inaccurate depth prediction. In this work, we consider the setting of camera-centric indoor environments with a few dozen input images. As the overlap between images is low, additional regularization is crucial. 

Thus, recent works \cite{dietnerf, pixelnerf, regnerf, dsnerf, densedepthprior, scade, sparsenerf} have enhanced NeRF when only sparse views are available. Many of these rely on data-driven priors or regularization to guide NeRF optimization. \cite{pixelnerf} equips NeRF with per-view image features from a CNN encoder that is trained on large multi-view datasets, enabling scene reconstruction from as little as one view in a single forward pass. \cite{dietnerf} supervises NeRF by enforcing consistency amongst CLIP \cite{clip} representations of arbitrary views. \cite{regnerf} samples unobserved views and regularizes the geometry and appearance of rendered patches from these views. 

\paragraph{NeRF with depth supervision:}
Amongst these, an interesting line of work explores supervising NeRF construction with depth \cite{densedepthprior, nerfingmvs, whatuncertainty, dsnerf, scade, sparsenerf, darf}. Notably, few of these works explicitly identify the inherent uncertainty in depth predictions. \cite{dsnerf} supervises NeRF ray sampling with COLMAP-derived \cite{colmap} depth and models noise with a Gaussian centered at the depth predictions. \cite{densedepthprior} train a custom network in-domain to predict uncertain areas. However, the model overfits to the training case, where high depth errors occur at the edges of objects. \cite{diner, sparsenerf} focus on object-centric applications with large overlap between train images. \cite{diner} employs predicted depth by using the difference between sample location and estimated depth as conditioning for the NeRF and for depth-guided ray sampling.  \cite{sparsenerf} supervises NeRF optimization by relaxing hard constraints in depth supervision to softer, more robust local depth ranking constraints. Although this approach acknowledges potential noise in pretrained depth network estimates, it does not directly incorporate uncertainty to guide depth supervision for NeRF. SCADE \cite{scade} leverages 20 ambiguity-aware depth proposals generated by a costly out-of-domain prior network and a space-carving loss with mode-seeking behavior to supervise ray termination distance in NeRF. We found that while some depth hypotheses can be accurate, many of these proposals may also be poor predictors of scene geometry. This finding challenges the utility of having multimodal depth proposals for resolving ambiguity (see Figure \ref{fig:multimodaldoesnthelp}), when NeRF naturally captures multimodal ray termination distributions. D\"{a}RF \cite{darf} is most closely related to our work and, like our work, proposes to use a standard pretrained depth estimation network to supervise NeRF training. Yet, D\"{a}RF does not consider pixel-wise uncertainty and supervises the NeRF-rendered depth directly instead of guiding the ray termination distance distribution.

\section{Method}
\label{sec:method}
\begin{figure*}
    \centering
    \includegraphics[width=\linewidth]{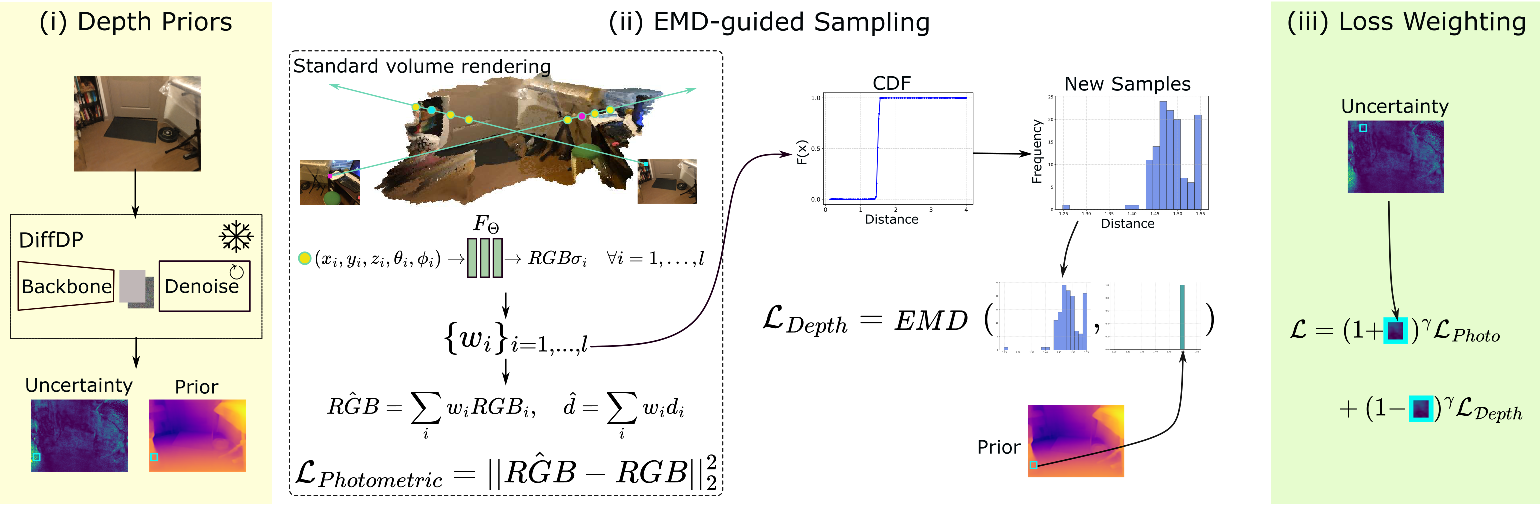}
    \caption{A detailed schematic of our depth-guided NeRF optimization. (i) A pretrained diffusion model for depth prediction, DiffDP \cite{ddp}, provides depth priors. Measuring the progression of depth predictions throughout the denoising process provides uncertainty maps. (ii) Given inputs poses ($x$,$y$,$z$,$\theta$,$\phi$) a network $F$ outputs RGB value and density. From the outputs, we derive weights $w$ that, when normalized, serve as a piece-wise-constant probability density function. We can then construct a cumulative distribution function (CDF) from which we sample new ray termination distances. We supervise these samples with the Earth Mover's Distance (EMD) to the depth prior. (iii) Finally, we weigh the photometric and depth losses according to the DiffDP-derived uncertainty.}
    \label{fig:archi}
\end{figure*}
An overview of our method is shown in Figure \ref{fig:archi}. In this section, we describe each of the three components of our pipeline (depth prior, EMD-guided ray termination sampling, and  photometric-geometric loss balancing) in turn.
\subsection{Background}
The core of our work is a standard neural radiance field (NeRF) \cite{nerf}. Given a collection of images $\{\mathbf{I}_i\}_{i=1,...,N}$ and their corresponding SfM camera poses, we aim to train a network to render images from novel, unseen views. Similar to existing works, we encode a NeRF as multilayer perceptron $F_{\Theta}: (x_i,y_i,z_i, \theta_{i}, \phi_{i} ) \rightarrow (\mathbf{c_{i}}, \sigma_{i})$ that takes camera position and viewing direction as input and outputs color $\mathbf{c}$ and volume density $\sigma$. To render depth from a given camera view, we cast a ray $\mathbf{r}$ through the origin $\mathbf{o}$ of the camera and a pixel projected into world space. We sample termination distances from the ray within a predefined interval $(\text{near}, \text{far})$ and pass them through a neural radiance field network that predicts their weights $w_i$ as probability that the ray can traverse space without obstructions until reaching the hypothesis in question:
$w_i = T_i(1-\exp(-\sigma_i\delta_i))$, 
where $\delta_i$ is the distance between samples, and $T_i = \exp(-\sum_{j=1}^{i-1}\sigma_j\delta_j)$. 
The final predicted color $\hat{C}$ and depth $\hat{{d}}$ is then the weighted average of all $l$ ray termination distances along $\mathbf{r}$
\begin{align}
\label{expected_depth}
   \hat{C}(\mathbf{r}) = \sum_l w_l\mathbf{c}_l \quad \text{and} \quad \hat{d}(\mathbf{r}) = \sum_l w_ld_l.
\end{align}

A standard NeRF is supervised with the photometric loss between observed color and the expected color
\begin{align}
    L_\text{photo} = || \hat{C}(\mathbf{r}) - {C}(\mathbf{r}) ||_2^2.
\end{align}
But in a setting with only $N<20$ images to learn from, and frequent occurence of large untextured areas such as white walls, a NeRF is easily underconstrained and does not learn the underlying geometry sufficiently. To improve geometric understanding, and to provide additional guidance during training, we incorporate depth supervision.   

\subsection{Depth Priors and Uncertainty}
As monocular depth estimation is ambiguous, we require a measure of trust in the predictions. While many networks display promising depth prediction capabilities, generative models, especially denoising diffusion models, additionally let us reason about the depth generation process.  DiffDP \cite{ddp} is a state-of-the-art image-conditioned diffusion model designed for visual perception tasks like dense segmentation and depth estimation. For depth estimation, the network is trained to denoise noised ground truth depth maps using image features as conditioning for the denoising process. At inference, multi-resolution features are extracted from the input image and concatenated with random noise, where a lightweight decoder gradually denoises the input to generate a final depth prediction. 
Given an input image $\mathbf{I}$, DiffDP formulates its denoising process as
\begin{align}
    p_{\theta}({z}_{0:T} | \mathbf{I})
 = p_{\theta}({z}_{T}) \prod_{t=1}^{T} p_{\theta}(z_{t-1}|z_t, \mathbf{I}),
 \end{align}
where $z_t \sim \mathcal{N}(0, {I})$ and $z_0$ corresponds to the models final depth estimate.

During this process, the network recursively updates its estimate. Pixels that the model is unsure about will be updated more often than others. We propose to use this measure as a proxy for uncertainty that is \textit{not} focused on areas of high error, such as borders of objects, but instead captures uncertainty in the depth generation process itself. 
 
\begin{figure}[t]
    \centering
    \includegraphics[width=0.7\linewidth]{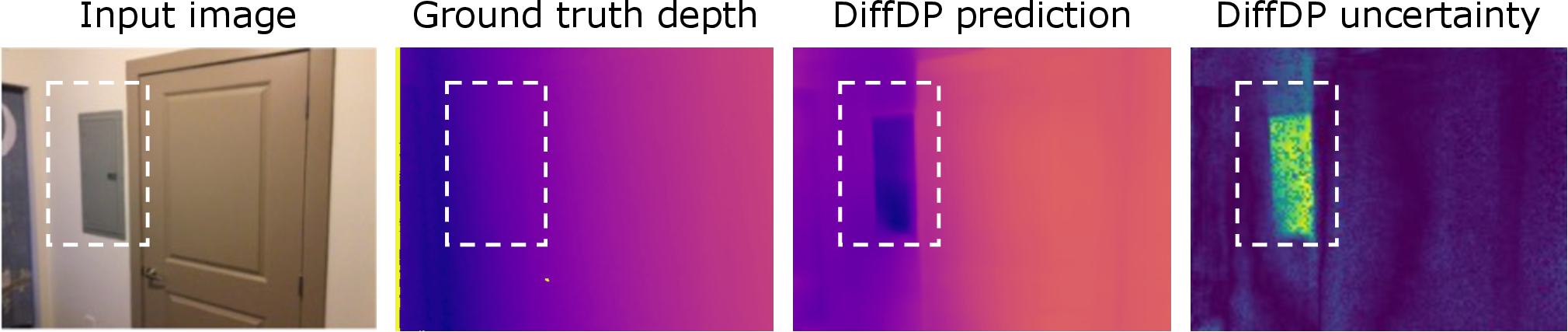}
    \caption{Example of DiffDP depth prediction that misinterprets the depicted geometry. Although the predicted depth has errors, the uncertainty map is able to highlight areas of large errors. This allows us to tune down the depth loss in unreliable areas.}
    \label{fig:ddp_bad}
\end{figure}

To obtain the uncertainty of a depth estimate, $u(z_0)$, we compare the current estimate to the previous one at each time step $t$ and report the count as:
\begin{align}
   c(z_{0:T}) =  \frac{1}{T} \sum_{t=T, ..., 1} \mathbbm{1}_{|z_t - z_{t-1}| \geq \tau}.
\end{align}
 Let $\mathcal{M}(\cdot)$ denote a function that mirrors an image, then the uncertainty for an image $x$ is
 \begin{align}
     U({z}_{0:T} | \mathbf{I}) = \frac{c(z_{0:T}|\mathbf{I}) + \mathcal{M}(c(z_{0:T}|\mathcal{M}(\mathbf{I})))}{2}.
 \end{align}
We empirically found this measure to find extreme cases of uncertainty easily, while emphasizing somewhat uncertain areas less. To capture those areas as well, we augment the uncertainty map with a second dimension of uncertainty: We compare the final depth prediction of an image with the prediction of its mirrored image. The final uncertainty for an image $I$ is then
\begin{align}
    u({z}_{0:T} | \mathbf{I}) = U({z}_{0:T} | \mathbf{I}) 
    \cdot ||(z_{0}|\mathbf{I}) - \mathcal{M}(z_0|\mathcal{M}(\mathbf{I}))||_1. 
\end{align}
An example uncertainty map is depicted in Figure \ref{fig:ddp_bad}, where DiffDP misinterprets the geometry of the electrical box, but highlights the same area as uncertain.

\subsection{$EMD$-Guided Sampling}
Our aim is to provide NeRF with depth \textit{suggestions}. Even though we use a state-of-the-art depth prior, and an uncertainty measure that provides information about unreliable depth priors, we should still not assume that the prior is a perfect pseudo ground truth. Minimizing a norm between the NeRF-rendered depth and the depth prior, such as proposed in the concurrent work D\"{a}RF, would force the NeRF to approximate the depth priors irrespective of their accuracy. This could not only lead to a wrong understanding of the underlying scene, but also interfere with the RGB loss, which propagates gradient updates through shared layers. We thus seek to guide the ray termination distance distribution with the depth prior instead of directly supervising the NeRF-rendered depth.

A second motivation to leverage the distance distribution stems from NeRF's design. The volume density $\sigma(y)$ represents the differential probability that a ray terminates at $y$. We found empirically, that the learned ray termination distance distribution is rarely unimodal. So if we supervised the expected depth in Eq. \ref{expected_depth} directly, we would collapse this probability into its expected value and lose this valuable information about the distribution. We thus supervise the ray termination distance distribution as proposed in SCADE \cite{scade} rather than supervising the predicted depth. To this end we normalize the predicted weights,$w_i$, to obtain a piecewise-constant probability density function along a ray: $\hat{w}_i = {w_i} / {\sum_{j=1}^{l}w_j}$.

We can then construct the cumulative density function (CDF) and sample new ray termination distances $\mathbf{y} = y_1, y_2, ...y_N$ through inverse transform sampling. The distribution of these samples can then be supervised in lieu of the less informative NeRF-sampled depth. But while SCADE models uncertainty in depth priors by providing 20 samples to choose from, the method still assumes that the chosen hypothesis is a perfect pseudo ground truth. We aim to relax this assumption.
We instead propose to supervise the ray termination distance distribution with the Earth Mover's distance between the NeRF samples and the depth prior $z_0$:
\begin{align}
    L_{\text{EMD}} = \text{EMD}(\mathbf{y}, z_0).
\end{align}

We use Earth Mover's distance, as it naturally lends itself to compare probability densities or discrete histograms. It does not require satisfaction of assumptions such as that for two distributions ${P}$ and ${Q}$, ${Q}(y) = 0 \implies {P}(y) = 0,\ \forall y$, which KL divergence would require. Earth Mover's distance also does not require that $y_1, y_2, ...y_N$ are unimodally distributed, or that the error between NeRF-sampled ray termination distances and depth prior is normally distributed. Earth Mover's distance is therefore equipped to address complex uncertainties in our depth priors, such as those we empirically observe and that previous methods have not sufficiently captured. As EMD is not differentiable, we use Sinkhorn Divergence \cite{sinkhorn} to approximate the $L_{\text{EMD}}$ during training.

\subsection{Loss Weighting}
\label{sec:loss_weighting}
Given the standard photometric loss and our novel depth guidance framework, we leverage the uncertainty we can capture for free from DiffDP's generation process to downweight pixels with especially uncertain depth prediction. Our work aims to provide a framework in which RGB-losses and depth losses complement each other: If we are certain about the depth of a pixel, we wish to upweight the depth loss. When we are unsure, we would rather rely on the photometric loss. Loosely inspired by Focal Loss \cite{lin2017focal}, we define the total loss for a ray as

\begin{align}
    \mathcal{L} = (1 + u)^\gamma L_{photo} + \lambda (1 - u)^\gamma L_{\text{EMD}},
\end{align}
where $\lambda$ is a balancing weight and $\gamma$ controls the impact of the uncertainties $u$. Opposed to Focal Loss which increases the weight of uncertain examples to force the model to learn hard cases, we apply less weight to uncertain pixels.

\section{Results}
\label{sec:results}
In this section we evaluate the proposed method in detail, compare it to existing work, and explore the importance of our design choices.

\subsection{Experimental Setup}
We evaluate our method on three ScanNet \cite{dai2017scannet} scenes as chosen by DDPrior, D\"{a}RF, and SCADE \cite{densedepthprior, scade, darf}. Each scene consists of 18-20 training images and 8 test images. To measure the generalization capability of our framework, we also evaluate on an additional dataset which we refer to as ScanNet+ containing additional scenes from ScanNet. Unless otherwise stated, all experiments are performed and averaged over the three standard ScanNet scenes. For evaluation, we use the standard photometric measures PSNR, SSIM, and LPIPS \cite{lpips} as used in the original NeRF paper \cite{nerf} and in all baselines we compare to. As we are especially interested in the ability to render accurate depths, we adapt the depth metrics used in D\"{a}RF, namely relative error (Rel) and RMSE \cite{darf}.

We compare our work to several baselines that use depth-supervision for NeRF training in indoor environments. The most related baseline is D\"{a}RF \cite{darf}, as it also leverage a frozen out-of-the-box pre-trained monocular depth prior to supervise NeRF training. SCADE is related to our work as well, but uses a custom-trained depth prior, and does not evaluate the depth accuracy of their method. As SCADE provides pre-trained weights, we evaluate their method on all depth metrics. Additional baselines are a standard NeRF \cite{nerf}, DS-NeRF \cite{dsnerf}, and DDPrior \cite{densedepthprior}, where we use the results reported by DDPrior. We report results on DDPrior using an out-of-domain prior as trained by SCADE, to replicate our setting where the depth network is trained on a different dataset. 

\subsection{Implementation Details}
\label{sec:implementation_details}
Our depth prior is the pretrained DiffDP network that was trained on the indoor depth dataset NYU \cite{couprie2013indoor}. During training we initialize the scale of the depth priors as $1$, and learn the scale with a small learning rate. The uncertainty maps from DiffDP are derived once during construction of the prior and used at the beginning of NeRF training (normalized to $[0,1]$). We use 1024 rays per batch, and sample $64$ and $128$ ray termination distances for the coarse and fine network, respectively. For the Earth Mover's distance we sample an additional $128$ samples. We use the \textit{geomloss} implementation of Sinkhorn with standard hyper-parameters. We employ dropout layers ($p=0.1$) and weight decay ($\lambda_{wd}=1e^{-6}$) for regularization. Further details of the prior construction, NeRF training, and evaluation can be found in the supplementary material.

\subsection{Reconstruction Quality}
\begin{figure}[t]
    \centering
    \includegraphics[width=0.8\linewidth]{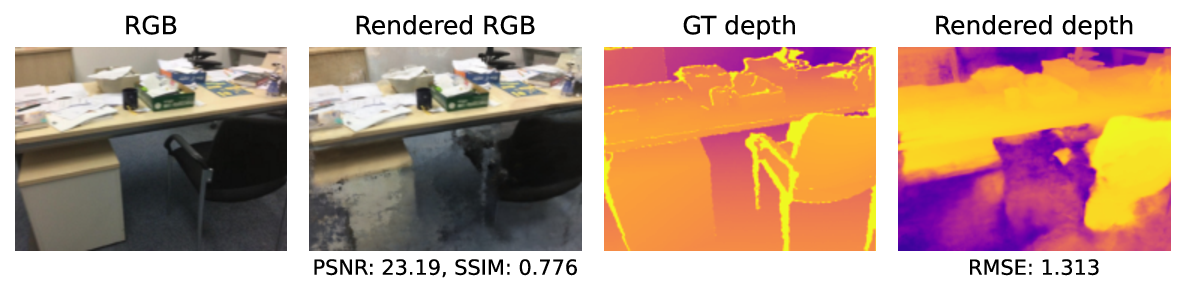}
    \caption{Good RGB rendering quality does not imply good geometric understanding. In this test example, SCADE \cite{scade} accurately renders the image (PSNR and SSIM above average), while misinterpreting the geometry of the depicted scene (five-fold average RMSE). The model does not capture the cabinet below the desk in the depth map.}
    \label{fig:rgbvsdepth}
\end{figure}
In this section we evaluate our method's ability to render novel views, and, importantly, its ability to understand the geometric layout of a scene. We thus report both photometric and depth-based metrics. But first, we demonstrate that photometric and depth errors do not always go hand in hand. Observe in Figure \ref{fig:rgbvsdepth}, that a model can yield great photometric results, while misunderstanding the geometry. In the illustrated example, the cabinet is visible in two training images only and from similar angles. This leads to a misinterpretation of the scene such that the carpet appears to have a beige box pattern. Correcting the geometry in such cases does not improve rendering quality.

\begin{table}[t]
\begin{center}
     \caption{Experimental results on ScanNet \cite{scannet}. D = Depth supervision during training: \ding{55} none, \ding{51} in-domain, \ding{66} out-of-domain.  DiffDP is the  depth prediction network we use as a prior. Our method dramatically improves NeRF's underlying scene geometry over initial depth priors while maintaing photometric reconstruction quality. Note that our method also outperforms DDPrior which has in-domain pretrained depth maps. }
    \resizebox{\linewidth}{!}{
    \begin{tabular}{l|c|c|c|c|c|c|c|c}
    \toprule
     \rowcolor{Gray!25} & & \multicolumn{3}{c}{RGB-based metrics} &\multicolumn{4}{|c}{Depth-based metrics} \\
    \midrule
 & \textbf{D} & PSNR $\uparrow$& SSIM $\uparrow$& LPIPS $\downarrow$& AbsRel $\downarrow$& SqRel $\downarrow$& RMSE $\downarrow$& RMSE log $\downarrow$\\
 \midrule
DiffDP \cite{ddp} & \ding{66}&-&-&-&0.100&0.032&0.261&0.128\\
  \midrule
NeRF \cite{nerf}&\ding{55}&19.03&0.670&0.398&-& - &1.163 &- \\
DS-NeRF \cite{dsnerf}&\ding{51}&20.85&0.713&0.344& -&-&0.447&-\\
DDPrior \cite{densedepthprior}&\ding{51} &20.96&0.737&0.294&-&-&0.236&-\\
DDPrior* \cite{densedepthprior}&\ding{66} &19.29&0.695&0.368&- &- & 0.474 &-\\
SCADE** \cite{scade}&\ding{66}&21.54&0.732&\textbf{0.292}&0.086&0.030&0.252&0.118\\
D\"{a}RF \cite{darf} &\ding{66}  &21.58&\textbf{0.765} &0.325&0.151&0.071&0.356&0.168\\ 
\midrule
\textbf{Ours} &\ding{66}&\textbf{21.69} &0.737&0.373&\textbf{0.070} &\textbf{0.024} &\textbf{0.221} &\textbf{0.105}\\  
\bottomrule
\multicolumn{9}{l}{\footnotesize{* Trained by SCADE authors. ** Depth metrics evaluated by us based on SCADE's weights.}}
 
    \end{tabular}}\\
    
    \label{tab:results_scannet}
    \end{center}
\end{table}
\begin{figure}[t]
    \centering
    \includegraphics[width=\linewidth]{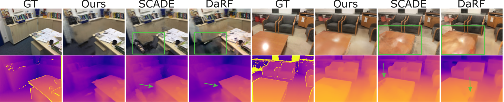}
    \caption{Qualitative results of rendered RGB images and depth maps. Our method produces less artifacts in the left example and learns a better geometric representation of the table in the right example. }
    \label{fig:qualitative_results}
\end{figure}
We compare our method on RGB-based and depth-based metrics in Table \ref{tab:results_scannet}. Our method reduces all depth metrics of all baselines by at least $11\%$, and reduces the error of the most related work, D\"{a}RF, by $54\%, 66\%, 38\%$, and $38\%$ on the four depth metrics. 
Our method also outperforms DDPrior \cite{densedepthprior} whose depth prior was trained in-domain. Our NeRF setup even improves the initial depth estimates of the depth prior DiffDP. 
Moreover, our method not only outperforms SCADE, but also avoids the need for a custom-trained depth prior. Training this prior requires evaluating the entire training set 20 times after each epoch to find the best hypothesis for each training example. Our training of SCADE's cIMLE-based prior takes more than one week on four 24GB NVIDIA Titan RTX GPUs.

When comparing RGB metrics to the closest baseline, SCADE, our method yields a $0.7\%$ worse SSIM, while the PSNR is $0.7\%$ better. Though our method yields an LPIPS that is on average worse than SCADE's reported result, when reproducing SCADE's results, some runs lead to an extremely high LPIPS above 0.5 (see Figure \ref{fig:seedissue}).
Compared to D\"{a}RF, our SSIM is $3.7\%$ worse, but all depth metrics are $37\%$ to $66\%$ improved. Overall, we highlight that when geometric consistency is desired, our method enables substantial gains in depth metrics while maintaining comparable RGB metrics. 

\begin{figure}[t]
    \centering
    \includegraphics[width=\linewidth]{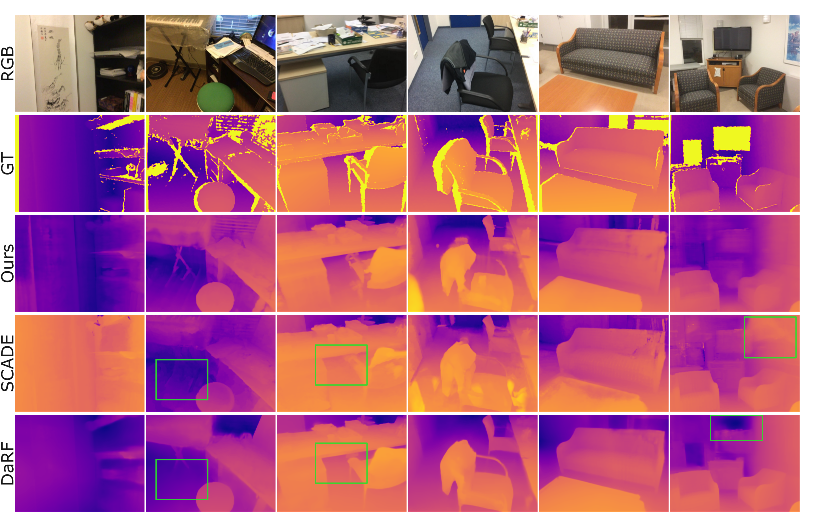}
    \caption{Additional qualitative depth results. Our method produces less artifacts than SCADE, while more accurately capturing the underlying geometry than D\"{a}RF. The smoothness of D\"{a}RF's predictions is a result of it's strong supervision by the depth prior, but does not speak to the accuracy of the learned geometry.}
    \label{fig:qualitative_results_depth}
\end{figure}

\indent We qualitatively compare results in Figure \ref{fig:qualitative_results}. In the illustrated examples, our method leads to less severe artifacts, and crisper and more accurate edges in RGB renderings and depth maps. Additional qualitative examples of depth maps are shown in Figure \ref{fig:qualitative_results_depth}. Our model more accurately reconstruct the depicted layouts, like the area underneath the piano in the second image, or the area under the table in the third image. Interestingly, D\"{a}RF's depths are extremely crisp. However, the quantitative evaluation in Table \ref{tab:results_scannet} confirms that D\"{a}RF learns an inaccurate representation of the geometry. Its strong $L_2$-loss enforces the depth maps to look like their prior, although they do not correctly represent the scene. 

To demonstrate the robustness of our method we report results on ScanNet+ that were not previously used by the baselines. We observe in Table \ref{tab:additional_scannet}, that our method reduces the RMSE of the baselines by at least $56\%$, while producing comparable photometric results.
Please see the supplement for more results.
\begin{table}
    \centering
    \caption{To further demonstrate the generalization strength of our approach, we evaluate on a second dataset, ScanNet+ , which includes novel scenes not tested by prior work. We retrain all prior methods on these scenes and compare with ours.}
    \resizebox{0.5\linewidth}{!}{
    \begin{tabular}{l|c|c|c|c|c}
    \toprule
    \rowcolor{Gray!25}& PSNR $\uparrow$& SSIM $\uparrow$& LPIPS $\downarrow$& AbsRel $\downarrow$& RMSE $\downarrow$\\
    \midrule
    SCADE \cite{scade} &20.57&0.687&0.369&0.396&1.032\\
    D\"{a}RF \cite{darf} &22.01&\textbf{0.751}&\textbf{0.319}&0.745&1.765\\
    \textbf{Ours}&\textbf{22.29}&0.722&0.391&\textbf{0.156} &\textbf{0.456} \\
    \bottomrule
    \end{tabular}
    }
    \label{tab:additional_scannet}
\end{table}

\subsection{Role of Depth Supervision and Uncertainty}
\label{sec:depthprior}

We wish to better understand how NeRFs leverage depth priors during training and conduct extensive ablations studies under various settings.

In Table \ref{tab:depth_prior_ablation} (a) we evaluate the importance of design choices of our model compared to a basline NeRF model. We ablate Regularization $Reg.$, use of depth supervision $z$ (by default with $L_2$ loss), use of $EMD$ depth supervision instead of $L_2$ loss, and uncertainty weighting $u$. We observe that depth supervision plays a vital role in improving results, especially depth metrics. But depth alone ($Reg.$ + $z$) does not yield satisfactory depth results. Replacing a naive $L_2$-loss with the EMD ($Reg.$ + $z$ + EMD + $u$) considerable reduces the depth error, highlighting the importance of depth guidance. 
\begin{table}[t]
    \centering
    \caption{Ablation studies for the components of our method, varying depth priors and the objectives leveraging them. All reported results are on ScanNet scenes. } 

    \begin{minipage}{0.49\textwidth}
        \resizebox{\columnwidth}{!}{
    \begin{tabular}{c|c|c|c|c|c|c|c|c}
    \toprule
        \rowcolor{Gray!25} Reg. & $z$ & $EMD$ & $u$ & PSNR $\uparrow$& SSIM $\uparrow$& LPIPS $\downarrow$& AbsRel $\downarrow$& RMSE $\downarrow$ \\
    \midrule
    \checkmark & & & & 21.28 & 0.735 & 0.370 & 0.119 & 0.354\\
    \checkmark & \checkmark & & & 21.24 & 0.734 & {0.377} & 0.100 & 0.295\\
    \checkmark & \checkmark & \checkmark & & \textbf{21.70} & 0.736 &{\textbf{0.369}} & 0.071 & 0.222\\ 
    \checkmark&\checkmark&\checkmark&\checkmark & 21.69 & \textbf{0.737} & 0.377 & \textbf{0.070}& \textbf{0.221}\\ 
    \bottomrule
    \end{tabular}} 
\scriptsize
        (a) Impact of Regularization $Reg.$, the DiffDP prior $z$, the Earth Mover's distance loss $EMD$, and uncertainty $u$ on the baseline NeRF model. 

        \resizebox{\columnwidth}{!}{
    \begin{tabular}{l|c|c|c|c}
    \toprule 
    \rowcolor{Gray!25} & AbsRel $\downarrow$& SqRel $\downarrow$& RMSE $\downarrow$& RMSE log $\downarrow$\\
    \midrule
    DiffDP &  \textbf{0.125}&\textbf{0.056}&\textbf{0.324}&\textbf{0.151}\\
    DepthAnything & 0.144 & 0.061 & 0.357 & 0.180 \\
    \bottomrule   
    \end{tabular}}

        \scriptsize
(b) Comparison of our depth prior DiffDP \cite{ddp} and DepthAnything \cite{depthanything} evaluated on the training images used as input to our model.
    \end{minipage}%
    \hfill
    \begin{minipage}{0.49\textwidth}
        \resizebox{\columnwidth}{!}{
    \begin{tabular}{l|c|c|c|c}
    \toprule
    \rowcolor{Gray!25} With our loss ($EMD$)& PSNR $\uparrow$& SSIM $\uparrow$& LPIPS $\downarrow$& RMSE $\downarrow$\\
    \midrule
    Ours w/ DiffDP\ddag & \textbf{21.70}&\textbf{0.736}&\textbf{0.369}&\textbf{0.222}\\
    Ours w/ DepthAnything\ddag& 21.62 & 0.732 & 0.378 & 0.290\\
    \bottomrule
    \end{tabular}}
    \scriptsize (c) Impact of the depth priors in (b) on our model. For both priors, our $EMD$ objective allows the NeRF to learn smaller depth errors than the priors used for supervision yield.

    \resizebox{\columnwidth}{!}{
    \begin{tabular}{l|c|c|c|c}
    \toprule
    \rowcolor{Gray!25} With $L_2$-H loss& PSNR $\uparrow$& SSIM $\uparrow$& LPIPS $\downarrow$& RMSE $\downarrow$\\
    \midrule
    Ours w/ DiffDP\ddag & \textbf{21.34} & \textbf{0.726} & \textbf{0.382} & \textbf{0.334} \\
    Ours w/ DepthAnything\ddag  &21.04 & 0.717 & 0.394 & 0.349\\
    \bottomrule
    \end{tabular}}
        \scriptsize (d) Comparison of different depth priors under a $L_2$-objective applied to depth hypotheses ($L_2$-H loss $\equiv$ space-carving loss \cite{scade} with only one depth
hypothesis). \ddag Without uncertainty $u$, as it cannot be used with DepthAnything prior.
    \end{minipage}

\label{tab:depth_prior_ablation}
\end{table}

\begin{figure}[t]
    \centering
    \begin{minipage}{0.49\textwidth}
    \centering
    \includegraphics[width=\linewidth]{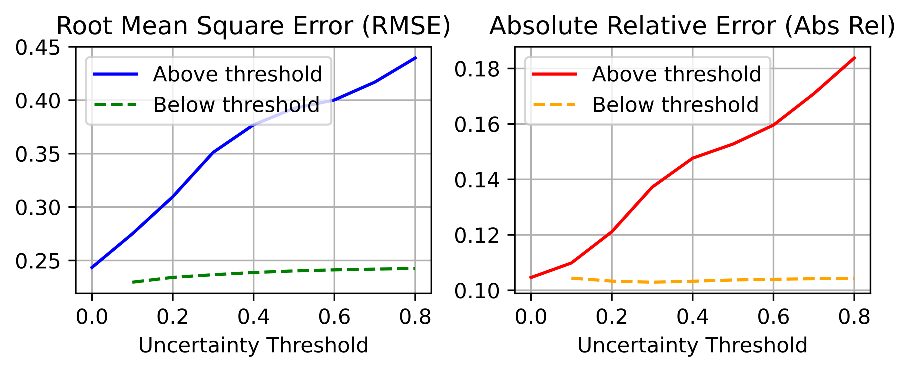}
    (a)
    \end{minipage}
    \begin{minipage}{0.49\textwidth}
    \centering
    \includegraphics[width=\linewidth]{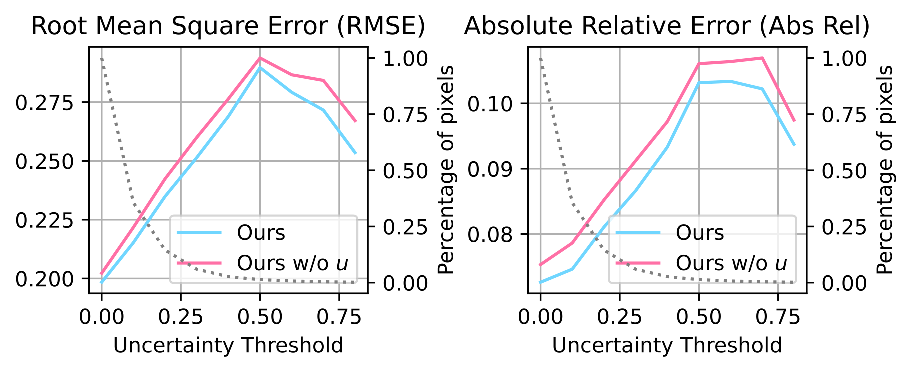}
    (b)
    \end{minipage}
        \caption{(a) Our uncertainty measure can identify areas in the DiffDP depth prior that we should not trust. We plot the error of the depth prior for pixels that are above or below a threshold for several thresholds. Both RMSE and Abs Rel are considerable higher in areas we predicted as uncertain. (b) We show the impact of incorporating our uncertainty $u$ in our model. We plot depth errors for all pixels above a uncertainty threshold and ablate our method in comparison to a version without $u$. Especially for high uncertainty ($>0.6$) incorporating $u$ helps improve depth accuracy.}
    \label{tab:uncertainty}
\end{figure}

Although the impact of $u$ is small when averaged across entire images, $u$ makes a difference where it matters: in uncertain areas. In Figure \ref{tab:uncertainty} (a), we evaluate how well the uncertainty measure can distinguish reliable from unreliable pixels. We compute the depth error of DiffDP for pixels that have an uncertainty of $\geq t$ and compare it to more certain pixels with  $ u < t$ for thresholds $t=\{0, 0.1, ... 0.8\}$. For $t=0.5$, for instance, the depth error of uncertain pixels is $63\%$ higher than the error of certain pixels. Our uncertainty measure thus reliably highlights pixels that should not be trusted. To evaluate its impact on our model, we ablate the use of $u$ in Figure \ref{tab:uncertainty} (b) in a similar manner as the ablation in Table \ref{tab:depth_prior_ablation} (a); but instead of averaging over an entire image, we look at uncertain and certain regions of an image. We observe, that most pixels have a low uncertainty (64\% of all pixels have an uncertainty of $<0.1$), and for those pixels incorporating uncertainty only marginally (1.9\%) helps. But for pixels with high uncertainty (>0.8), incorporating $u$ improves the RMSE by 5.1\%. 

\begin{figure}
\centering
\includegraphics[width=\linewidth]{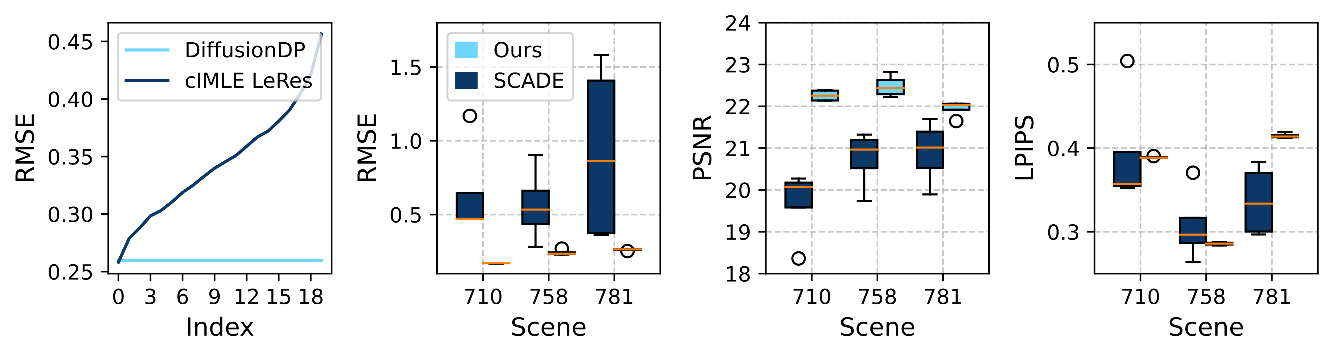}
\begin{minipage}{0.24\textwidth}
    \centering

    \caption{Depth errors of cIMLE priors increase rapidly when arranged in ascending order.}
    \label{fig:multimodaldoesnthelp}
\end{minipage} \hspace{8pt}
\begin{minipage}{0.72\textwidth}
    \centering
    
    \caption{We repeat the experiment in Table \ref{tab:results_scannet} with four random seeds and show the depth RMSE, PSNR, and LPIPS for each scene in ScanNet. While SCADE can sometimes produce accurate depths (low RMSE), the variability between runs is large. Our method produces consistently accurate depths.  }\label{fig:seedissue}
\end{minipage}
\end{figure}

It is important to note that our model's ability to predict accurate depth does not solely stem from a good prior.  We compare our prior to SCADE's LeReS-based prior \cite{leres} in Figure \ref{fig:multimodaldoesnthelp}, where we adjusted the scale between prior and ground truth as ratio between their mean depths before calculating the RMSE. We make two interesting observations.
First, sorting SCADE's 20 priors by accuracy, we find that the error increases rapidly. The additional priors introduce noise rather than adding beneficial information. This might explain why SCADE's training is at times unstable, and why learning accurate depth is so important for NeRFs. We can observe in Figure \ref{fig:seedissue} that SCADE depends highly on the chosen random seed. The depth RMSE varies dramatically between runs. Comparing this to the PSNR, we observe that even when the depth prediction is widely off (depth error of 1.6 in scene 781), the NeRF can still produce decent RGB quality with a PSNR of at least 20. So while a NeRF can learn to produce realistic looking RGB images, it has no knowledge of the geometry (as seen in Figure \ref{fig:rgbvsdepth}). This not only prevents applications such as augmented reality in which a user would interact with the 3D world or when meshes need to be extracted for the scene, but also renders reported photometric results unreliable. 

In Table \ref{tab:depth_prior_ablation} (b), we evaluate the initial accuracy of our DiffDP based prior and the prior from \cite{depthanything}. We show that DiffDP is a good choice for depth prior even in comparison to priors generated from large-scale pretraining like in DepthAnything \cite{depthanything}. Given this context, Table \ref{tab:depth_prior_ablation} (c) exemplifies our $EMD$-based objective's role in leveraging information from the priors. Irrespective of prior, our loss encourages a reduction in the initial prior's depth error (with our DiffDP prior, RMSE drops 31.4\%, and with DepthAnything prior, RMSE drops 18.8\%). Through $EMD$ depth guidance, NeRFs can leverage these priors without enforcing them, leading to a smaller overall depth error. To further examine the impact of our objective, we construct a strong baseline under different depth priors that mimics the loss presented in \cite{scade} but with a single depth hypothesis. Specifically, we use $L_2$ loss on depth hypotheses during NeRF training. We make two observations in Table \ref{tab:depth_prior_ablation} (d): First, depth metrics remain constant or even worsen in comparison to the initial depth priors from (b) under the $L_2$ objective. Secondly, our $EMD$-loss strongly outperforms the $L_2$ objective for both priors. 
These experiments highlight the importance of supervising NeRF with an EMD-based objective that can selectively incorporate information from the depth prior irrespective of the prior's strength.

\section{Conclusions}
While NeRFs can reliably render images from novel viewpoints, the underlying geometry is not always accurately learned. To improve the geometric understanding of NeRFs, we revisit depth supervision in NeRF training for the reconstruction of challenging indoor scenes. We present a simple, novel framework that leverages the depth estimates from pretrained diffusion models and their intrinsic notions of depth uncertainty. We show that noisy depth estimates coming from off-the-shelf depth estimation networks should not be used to directly supervise NeRF-rendered depths. Rather, weighted by uncertainty, the distribution of ray termination distances during NeRF optimization should be guided by depth priors through the Earth Mover's distance, allowing for selective supervision. Our method achieves strong empirical results and serves as an easy drop-in replacement for existing depth-supervised NeRFs. 
\paragraph{Limitations and Future work:} While different monocular depth priors can be leveraged in our $EMD$-guided framework, our uncertainty measure can only be obtained from diffusion-based depth networks. Future work should therefore include model-agnostic uncertainty measures. Additionally, the construction of the uncertainty map for the depth estimates can be sensitive to hyperparameters. An interesting direction for future work could be dynamically learning a threshold of uncertainty during the construction of the depth prior. 

\section*{Acknowledgements}
Thanks to the anonymous reviewers for their constructive feedback. This work was supported by the Isackson Family Foundation, the Stanford Head and Neck Surgery Research Fund, and the Stanford Graduate Fellowship.
%
%
\bibliographystyle{splncs04}
\bibliography{main}
\newpage
\appendix
\input{supplement}
\end{document}

%% file: supplement.tex
\clearpage
\setcounter{page}{1}
\setcounter{table}{0}
\renewcommand{\thetable}{S\arabic{table}}
\setcounter{figure}{0}
\renewcommand{\thefigure}{S\arabic{figure}}

\title{Depth-guided NeRF Training via Earth Mover’s Distance: Supplementary Material} 

\titlerunning{Depth-guided NeRF Training via Earth Mover’s Distance}

\author{
Anita Rau\and
Josiah Aklilu\and
F. Christopher Holsinger \and
Serena Yeung-Levy
}

\authorrunning{A.~Rau et al.}

\institute{
Stanford University \\
\email{\{arau, josaklil, holsinger, syyeung\}@stanford.edu} \\
\url{https://anitarau.github.io/emd-nerf.github.io/}
}

\maketitle

In this supplementary material we report additional qualitative results, secondary ablation studies, and additional implementation details. We also test the limits of our method by evaluating it on outdoor environments. 

\section{Additional Qualitative Results}
\begin{figure}
    \centering
    \includegraphics[width=\linewidth]{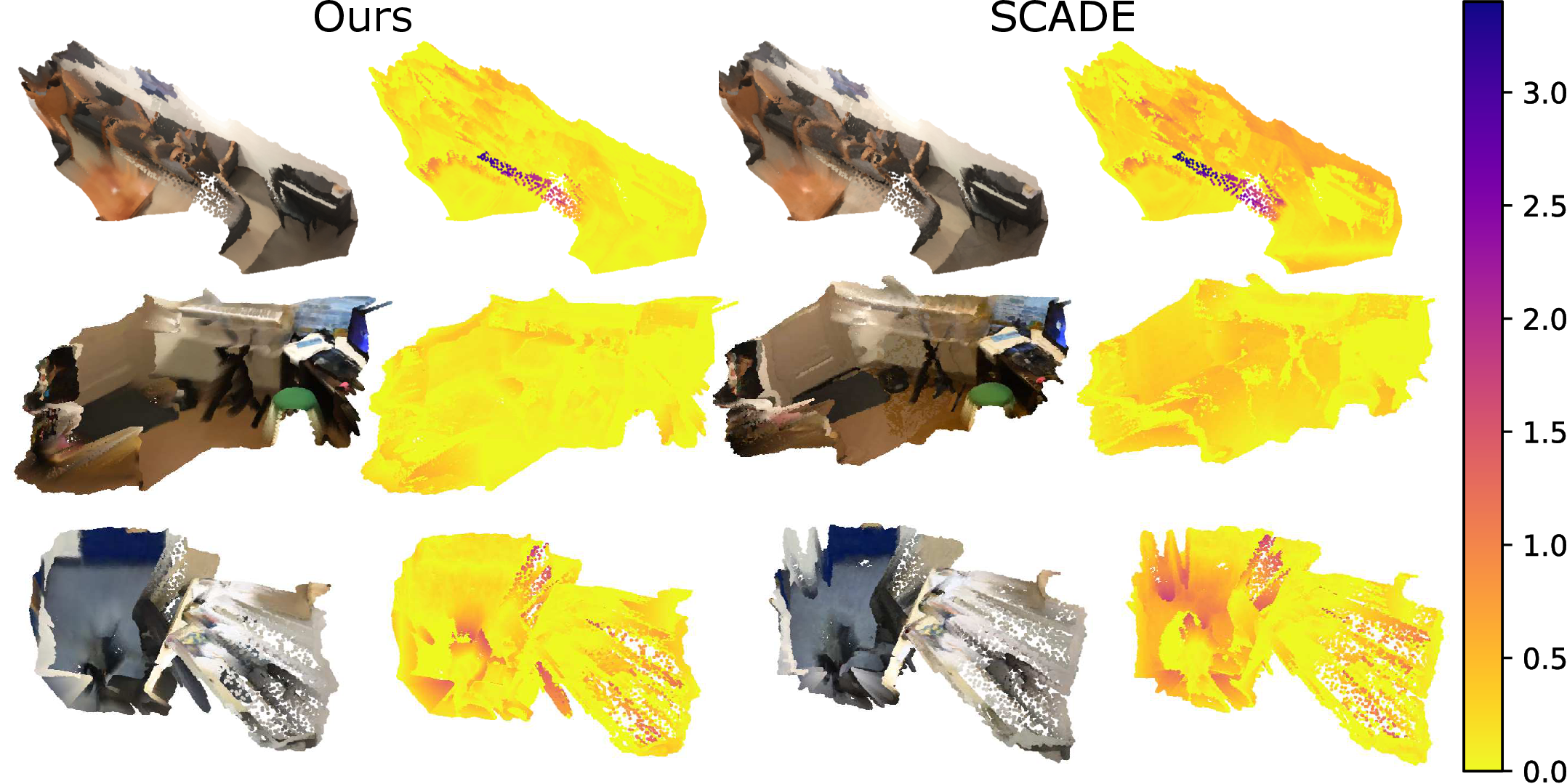}
    \caption{Comparison of the learned geometry in 3D. Each plot encompasses the concatenated 3D clouds of two test images with small overlap of each scene in ScanNet. Depth and RGB values are rendered. Error maps in m. Invalid ground truth values are set to zero (yellow).}
    \label{fig:clouds}
\end{figure}

To supplement Section 4.3 in the manuscript, we report additional qualitative results. We project the learned geometry and the rendered RGB values into 3D in Figure \ref{fig:clouds}. In each example, we show two test images from a scene concatenated into a mutual 3D cloud with some but minimal overlap between the images. We also project the depth errors onto the 3D structure, where darker colors indicate higher errors. 

\begin{figure*}
    \centering
    \includegraphics[width=\textwidth]{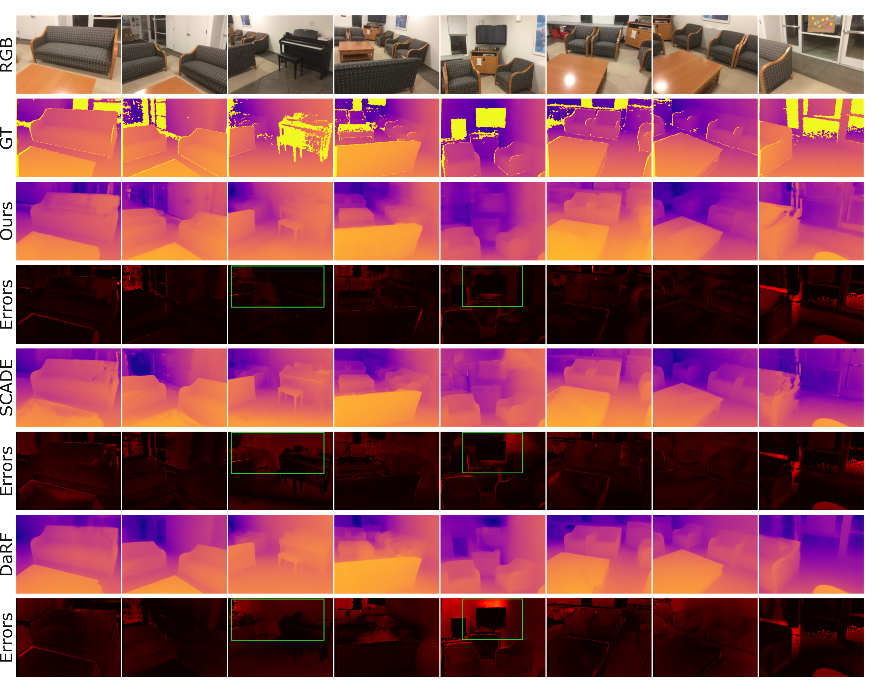}
    \caption{Qualitative comparison of our method with SCADE and D\"{a}RF. Our method yields more accurate NeRF-rendered depths. We depict the rendered depth maps and the corresponding error maps for all test images in ScanNet scene 781, where red corresponds to a higher error. All error maps in a column have the same scale, so colors are directly comparable. The baselines SCADE and D\"{a}RF produce brighter error maps speaking to a higher depth error. Although D\"{a}RF's depth maps look very crisp, they are geometrically inaccurate as, for instance, highlighted by the green boxes, where our method yields darker (thus more accurate) error maps. We masked out errors with invalid ground truth depth (yellow pixels in the ground truth GT).}
    \label{fig:err_maps}
\end{figure*}

Figure \ref{fig:err_maps} depicts depth predictions and corresponding error maps for all test images in one of ScanNet's scenes \cite{scannet}. Similar to the results in Figure 5, we observe more artifacts in SCADE's \cite{scade} predictions. D\"{a}RF's \cite{darf} depth looks promising but leads to a high depth error, manifesting in a brighter red color of the error maps compared to our method. These findings are consistent with the quantitative results in Table 1.

\begin{figure}
    \centering
    \includegraphics[width=\textwidth,trim={0 10.5cm 0 0},clip]{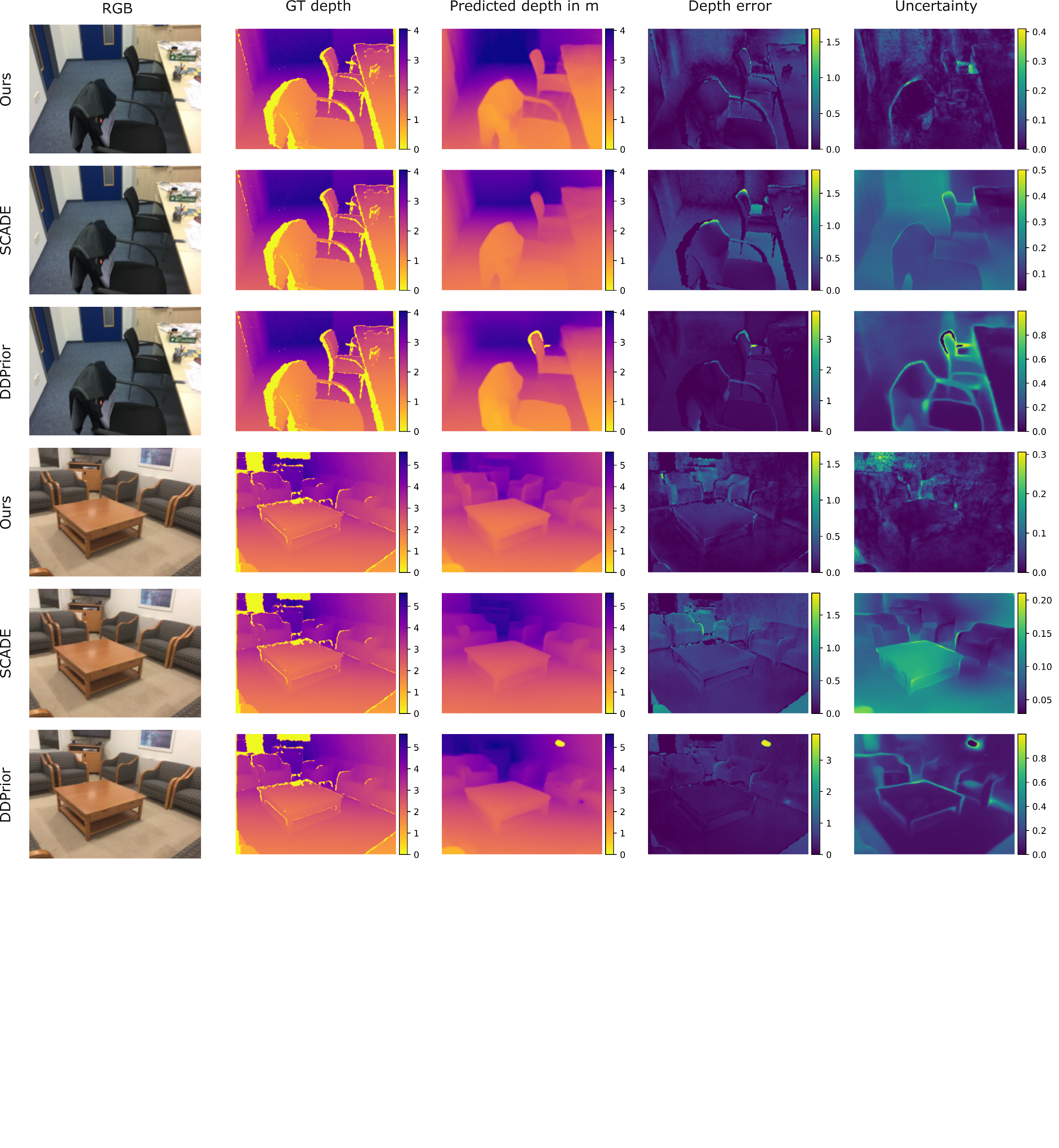}
    \caption{Comparison of different depth priors and their induced uncertainty maps. Ideally, the uncertainty should be high (yellow) around areas with a high depth error (yellow). Such behavior indicates that the uncertainty measure is a valuable tool to distinguish between those areas of an image whose depth prior we should trust and those we should not. However, we can observe that SCADE's uncertainty is high in most places, regardless of whether or not the depth prediction is accurate. DDPrior's uncertainty is high around edges, mirroring the training procedure of DDPrior, where high uncertainty implicates pixels with high depth loss. During training, these usually occur around edges, as the model overfits to a geometry. However, at test time, large depth errors occur mainly due to a misinterpreted shape of a scene, not due to steps in depth. Note also that DDPrior was trained in-domain. Our method instead highlights regions as uncertainty that indeed lead to the highest depth error.}
    \label{fig:unc_overview}
\end{figure}
Figure \ref{fig:unc_overview} illustrates the different concepts of uncertainty that different methods employ. SCADE predicts 20 depth hypotheses for each input image. We depict the standard deviation over the 20 hypotheses to illustrate where the hypotheses differ the most and, thus, where the prior is uncertain. DDPrior trains a depth network to predict uncertainty along with depth as an additional output channel \cite{densedepthprior}. For both SCADE and DDPrior, we use author-provided model weights.  D\"{a}RF models confidence as binary masks indicating whether the difference between the rendered depth and the depth hypothesis is below a threshold. As this confidence measure is not induced by the depth prior, but is derived in comparison to the rendered depth, we do not report it here. Further, D\"{a}RF's confidence changes throughout training and is applied to random patches only.

A good measure of uncertainty identifies which parts of a depth prior are unreliable. In Figure \ref{fig:unc_overview}, we observe that SCADE's depth prior is uncertain in most of the images and does not distinguish between pixels with high depth error and low depth error. DDPrior highlights pixels around edges the most due to its pretraining strategy. However, when applied out-of-domain at test time, the highest depth error does not occur around edges. The benefit of the uncertainty is, therefore, limited. Our uncertainty highlights the regions that lead to the highest depth error. Such errors are unavoidable when testing a depth prediction network out-of-domain. However, the ability to predict where the errors are high allows us to use depth priors more effectively. 

\section{Additional Ablation Studies}

Further to Section 4.4 in the main paper, we report additional ablation studies. We focus on the baselines SCADE and D\"{a}RF, as they are most closely related to our work and yield the most competitive results to ours. 

We wish to add insights into the importance of the chosen depth prior. To show that {DiffDP} \cite{ddp} alone is insufficient to improve the baseline methods, we swap in our prior for the baseline's depth priors in SCADE and D\"{a}RF. Using the baselines' implementations and keeping everything else constant, we report results in Table \ref{tab:prior_ablation}. 
\begin{table}
    \centering
    \caption{Baselines using our DiffDP prior. Our prior does not impact the predictions of the baselines notably. We conclude that improvements of our method versus the baselines do not stem from the choice of prior alone. We retrained D\"{a}RF to allow direct comparison with \textit{D\"{a}RF w/ DiffDP} as no official depth evaluation script exists.}
    \resizebox{0.75\columnwidth}{!}{
    \begin{tabular}{l|c|c|c|c|c}
    \toprule
    \rowcolor{Gray!15}  & PSNR $\uparrow$& SSIM $\uparrow$& LPIPS $\downarrow$& AbsRel $\downarrow$&  RMSE $\downarrow$ \\
 \midrule
\textbf{Ours w/ DiffDP (original)}&\textbf{21.69} &0.737&0.373&\textbf{0.070} &\textbf{0.221} \\  
 \hline
SCADE w/ DiffDP& 21.51 & 0.732 & 0.376 & 0.091 & 0.319\\ 
SCADE w/ LeRes cIMLE (original) & 21.54&0.732&\textbf{0.292}&0.086&0.252\\
 \hline
D\"{a}RF w/ DiffDP& 21.19&0.762&0.324&0.086&0.279  \\
D\"{a}RF w/ MiDaS (original, retrained) & 21.55&\textbf{{0.766}} &0.319&0.085&0.259\\
    \bottomrule
    \end{tabular}}
    \label{tab:prior_ablation}
\end{table}

As we find the depth evaluation protocol of  D\"{a}RF ambiguous, we retrain D\"{a}RF using the official implementation and report our evaluation metrics. In D\"{a}RF's case, DiffDP leads to slightly worse results across all metrics. For SCADE, DiffDP prior yields worse depth results than SCADE's multi-modal prior but slightly better photometric results as measured by PSNR and SSIM. We conclude that our depth prior alone is not sufficient to improve baselines.

\begin{figure}
    \centering
    \includegraphics[width=\columnwidth]{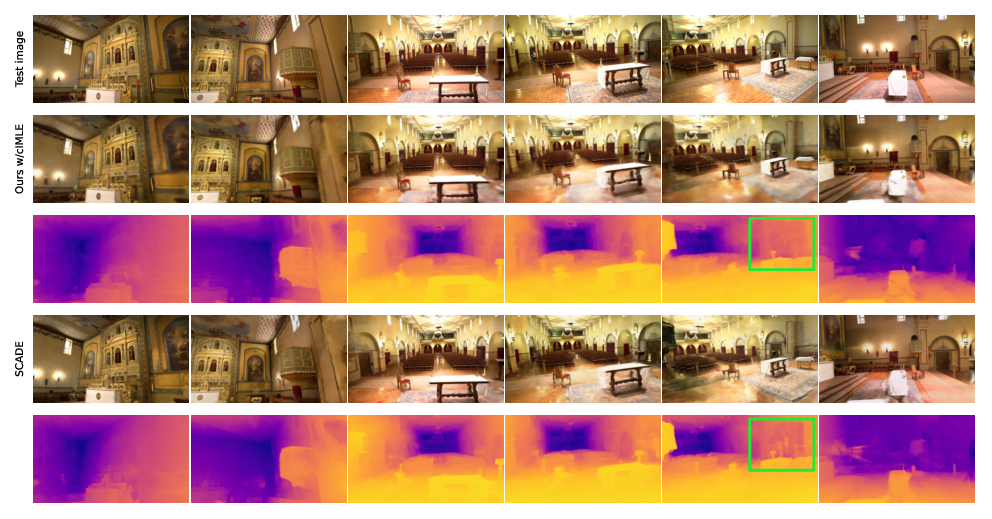}
    \caption{Qualitative results on indoor TnT scene \textit{Church}. Note that some geometrical details are better preserved by our method (green box), although the same depth prior is used.}
    \label{fig:tnt}
\end{figure}

\begin{table}
     \caption{Comparison of SCADE's LeRes cIMLE prior on SCADE's Tanks and Temples scenes.}
    \resizebox{\linewidth}{!}{
    \begin{tabular}{l|c|c|c|c|c|c|c}
    \toprule
     \rowcolor{Gray!25} &  \multicolumn{3}{c}{RGB-based metrics} &\multicolumn{4}{|c}{Depth-based metrics} \\
    \midrule
& PSNR $\uparrow$& SSIM $\uparrow$& LPIPS $\downarrow$& AbsRel $\downarrow$& SqRel $\downarrow$& RMSE $\downarrow$& RMSE log $\downarrow$\\
 \midrule

{Ours w/ LeRes cIMLE } &\textbf{20.23}&0.654&0.457&\textbf{0.097}&\textbf{0.159}&\textbf{0.915}&\textbf{0.194}\\ 
SCADE* w/ LeRes cIMLE (original)&19.83&\textbf{0.664}&\textbf{0.347}&0.098&0.171&0.937&0.213\\

\bottomrule

    \end{tabular}}\\
   \scriptsize * Trained by SCADE authors, evaluated based on official weights and evaluation script.
    \label{tab:indoor_tnt}
\end{table}
Conversely, we investigated whether our framework can leverage SCADE's prior in indoor environments. We are especially interested in out-of-distribution scenes to evaluate situations in which the depth prior is uncertain.  In Table \ref{tab:indoor_tnt} we compare the use of the LeRes cIMLE prior \cite{scade} in both SCADE and our method on the indoor Tanks and Temples scenes used by SCADE. Some qualitative results are shown in Figure \ref{fig:tnt}. For all Tanks and Temples experiments, we initialize the learnable scale parameter as mean over all ratios between sparse SfM estimates and depth prior as proposed by SCADE. We can observe that our method produces smaller depth errors using the same priors as SCADE. As the Tanks and Temples scenes differ from standard scenes used to pretrained the depth prior (a church, a courtroom, and an auditorium), we hypothesize that our method leveraging Earth Mover's Distance could be better equipped to capture the uncertainty in the depth priors. However, these results should be interpreted with care, as we only have information from very sparse SfM ground truth typically located around edges.  Note that during training we use all 20 depth hypothesis to compute the $EMD$ loss.

Lastly, for completeness, in addition to comparing the depth priors DiffDP and DepthAnything under $EMD$ and $L_2$-H losses in Tables 3(c) and 3(d), we compare the priors under the standard $L_2$ loss in Table \ref{tab:l2}. Unsurprisingly, DiffDP, the more accurate prior (see Table 3(a)) leads to a smaller depth error.

\begin{table}
    \centering
    \caption{Loss ablation study for depth priors DiffDP and DepthAnything. In addition to $EMD$ and space carving loss in Table 3(c,d), we add results using $L_2$ loss.}
    \resizebox{0.49\columnwidth}{!}{
    \begin{tabular}{l|c|c|c|c}
    \toprule
    \rowcolor{Gray!25} $L_2$ loss & PSNR $\uparrow$& SSIM $\uparrow$& LPIPS $\downarrow$& RMSE $\downarrow$\\
    \midrule
    Ours w/ DiffDP\ddag & \textbf{{21.24}} & {0.734} & {0.377} & \textbf{{0.295}} \\
    Ours w/ DepthAnything\ddag  &21.14 & \textbf{{0.737}} & \textbf{{0.364}}  & 0.353\\
    \bottomrule
    \end{tabular}}
    
        \tiny{\ddag Without uncertainty $u$, as it cannot be used with DepthAnything prior.}

    \label{tab:l2}
\end{table}

\section{Outdoor Experiments}
Although our method is specifically for indoor environments, we demonstrate generalizability to outdoor scenes using our original DiffDP prior pretrained on indoor NYUv2 data \cite{couprie2013indoor}. We report results on outdoor Tanks and Temples scenes used in D\"{a}RF. As D\"{a}RF did not publish the reconstructed scenes, we ran our own COLMAP pipeline. Note that a quantitative comparison to the D\"{a}RF paper based on two different sparse COLMAP reconstructions is not meaningful, as the overall scale of the scenes will be different. To enable future comparisons, we will make our COLMAP reconstructions of the outdoor scenes \textit{Truck} and \textit{Ignatius} available. We do not compare to SCADE, as their prior is trained on indoor scenes only.
Note that no dense ground truth depth exists, and the depth evaluation is based on the sparse COLMAP reconstruction. As the reconstruction is extremely sparse, depth results are to be interpreted with care. 

\begin{figure}
    \centering
    \includegraphics[width=\textwidth]{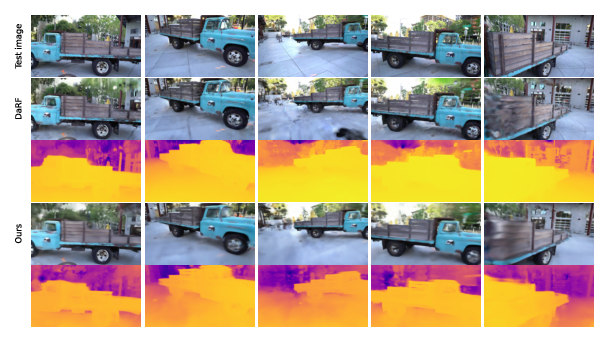}
    \caption{Rendered RGB images and depths on the Tanks and Temples \textit{Truck} scene.}
    \label{fig:truck}
\end{figure}

\begin{figure}
    \centering
    \includegraphics[width=\textwidth]{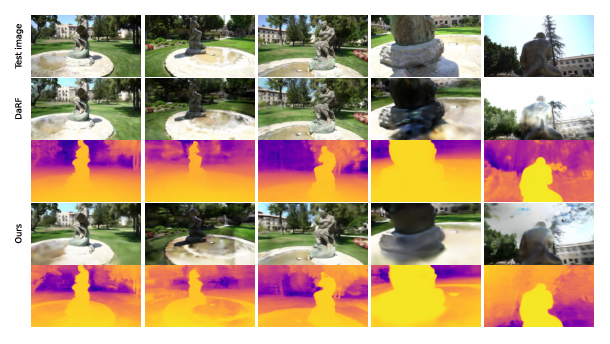}
    \caption{Rendered RGB images and depths on the Tanks and Temples \textit{Ignatius} scene.}
    \label{fig:ignatius}
\end{figure}

Although our DiffDP prior was trained on indoor scenes, while D\"{a}RF's MiDaS prior was trained on indoor and outdoor data, our method outperforms D\"{a}RF on both depth metrics in Table \ref{tab:outdoor_tnt}. To fully elucidate the influence of priors versus the model design, we additionally trained both methods using the same DepthAnything depth priors obtained from the official outdoor metric weights \cite{depthanything}. We can observe that even when both models leverage the same prior, our model outperforms D\"{a}RF on the depth metrics. Interestingly, our prior, although trained indoor, leads to the best geometric results. One reason might be that DiffDP with our uncertainty measure can compensate for uncertainty in outdoor environments, such as sky or far away surfaces. We show qualitative results for both scenes in Figures \ref{fig:truck} and \ref{fig:ignatius}. Similar to the results on ScanNet in Figure 6 of the main paper, D\"{a}RF's depth maps look crisp. This does not speak to the accuracy of the rendered depth but to the strict enforcement of the MiDaS prior. Rendered depths may look well while leading to high errors (see also Figure \ref{fig:err_maps} for further examples of this behavior).

\begin{table}
    \centering
    \caption{Comparison of our method with D\"{a}RF averaged over outdoor Tank and Temples scenes \textit{Truck} and \textit{Ignatius}.}
    \resizebox{0.75\columnwidth}{!}{
    \begin{tabular}{l|c|c|c|c|c}
    \toprule
    \rowcolor{Gray!15}  & PSNR $\uparrow$& SSIM $\uparrow$& LPIPS $\downarrow$& AbsRel $\downarrow$&  RMSE $\downarrow$ \\
 \midrule
{Ours w/ {DiffDP}} (original) & \textbf{16.39} & 0.462 & 0.692 &  \textbf{0.216 }& \textbf{ 1.539}  \\
Ours with DepthAnything & 16.05 & 0.457 & 0.701 &  0.248 &   2.193  \\
D\"{a}RF w/ MiDaS (original) &16.14 &\textbf{0.497}&\textbf{0.668}& 0.580&2.362 \\
D\"{a}RF w/ DepthAnything & 16.14 & 0.499 & 0.669 & 0.579 & 2.361 \\
\bottomrule
    \end{tabular}}
    \label{tab:outdoor_tnt}
\end{table}
\section{Additional Implementation Details}
In addition to Section 4.2 in the manuscript, this section provides further implementation and evaluation details.
\subsection{Our method}
\paragraph{Depth prior} Although {DiffDP} \cite{ddp} is a generative model, there is little variation between samples from the frozen network. The standard deviation for a pixel is usually in the range of around 1e-5. We thus used a very small threshold, $\tau$, to evaluate the uncertainty of a depth map. We obtained the threshold as 
\begin{align}
    \tau &= (\text{max}\textunderscore\text{depth} - \text{min}\textunderscore\text{depth})  \cdot 0.0001,
\end{align}
where max\textunderscore depth $= 10$ and min\textunderscore depth $= 0.001$ for ScanNet.

\paragraph{Training} Our network architecture builds upon the original NeRF \cite{nerf}. We added dropout layers with a dropout probability of $0.1$ after each of the first eight fully connected ReLU-activated layers that encode the positional encoding. We used a depth loss weight of $\lambda=0.007$ for all scenes except $710$, where we used $0.02$. The uncertainty weight $\gamma$ was $1$ in all of our experiments. We used Adam Optimizer with a learning rate of $5e^{-4}$, which we reduced to $5e^{-5}$ after $80\%$ of steps. We added a weight decay of $1e^{-6}$ and trained the model for a total of $500,000$ steps. We used a learning rate of $1e^{-7}$ for the depth scale.
\paragraph{Evaluation}
Let $1,...,K$ be the indices of pixels in an image that have a valid ground truth depth. Then the depth evaluation metrics between ground truth depth $d$ and predicted depth $\hat{d}$ are computed as follows:
\begin{align}
    \text{RMSE}(d, \hat{d}) &= \sqrt{\frac{1}{K} \sum_{k=1,...,K}(d_k - \hat{d}_k)^2}\\
    \text{RMSE}~log(d, \hat{d}) &= \sqrt{\frac{1}{K} \sum_{k=1,...,K}(\log{d_k} - \log{\hat{d}_k})^2} \\
    \text{AbsRel}(d, \hat{d}) &= \frac{1}{K} \sum_{k=1,...,K}\dfrac{||d_k - \hat{d}_k||_1}{d_k} \\
        \text{SqRel}(d, \hat{d}) &= \frac{1}{K} \sum_{k=1,...,K}\dfrac{(d_k - \hat{d}_k)^2}{d_k}
\end{align}
We report the mean over all depth maps in the test set.

\subsection{D\"{a}RF details}

For results in Table 2 in the main manuscript, we use the official D\"{a}RF implementation\footnote{\url{https://github.com/KU-CVLAB/DaRF}}. As pretrained model weights are not available, we retrained D\"{a}RF on ScanNet to provide the qualitative results in Figures 5 and 6 in the main manuscript. Our retrained version of D\"{a}RF achieved PSNR, SSIM, and LPIPS of 21.55,	0.766, and	0.319, which closely reproduces the author-reported results. Using D\"{a}RF's evaluation function, we obtain unusually high depth metrics of 0.682 and 1.613 for the AbsRel and the RMSE, respectively. However, we observed that D\"{a}RF  rescales the COLMAP-estimated input poses during training. Rescaling the predicted depth maps by the provided factors, we obtained depth AbsRel, SqRel, RMSE, and RMSE log of 0.085, 0.030, 0.259, and 0.106 respectively. Note that these values are considerably better than the author-reported results in Table 1. We did not receive confirmation about the exact details of D\"{a}RF's evaluation. We therefore reported qualitative results without absolute scales in Figures 5 and 6.

We use the rescaled depth maps (better than author-reported) for comparisons in Figures \ref{fig:err_maps} and \ref{fig:unc_overview}. For the results in Table 2, we assume a scale of 1, since the authors do not report details on how they obtained scales used in their code base.